\documentclass[10pt,twocolumn,letterpaper]{article}

\usepackage{cvpr}
\usepackage{times}
\usepackage{epsfig}
\usepackage{graphicx}
\usepackage{amsmath}
\usepackage{amssymb}
\usepackage{multirow}
\usepackage{subfigure}
\usepackage{pifont}
\usepackage{xcolor}
\usepackage{array}
\usepackage{url}
\usepackage{makecell}

\newcommand{\tabincell}[2]{\begin{tabular}{@{}#1@{}}#2\end{tabular}}

\newcolumntype{I}{!{\vrule width 1.2pt}}
\newlength\savedwidth

\cvprfinalcopy % *** Uncomment this line for the final submission

\ifcvprfinal\fi
\begin{document}
	
	%%%%%%%%% TITLE
\title{C$^3$ Framework: An Open-source PyTorch Code for Crowd Counting}

\author{Junyu Gao$^1$, Wei Lin$^1$, Bin Zhao$^1$, Dong Wang$^1$, Chenyu Gao$^1$, Jun Wen$^2$\\
	$^1$Northwestern Polytechnical University, Xi'an, Shaanxi, P. R. China\\
	$^2$College  of   Computer  Science  and  Technology,  Zhejiang  University, Hangzhou, Zhejiang, P. R. China\\
	{\tt\small \{gjy3035, elonlin24, binzhao111, nwpuwangdong\}@gmail.com,} \\
	{\tt\small  chenyugao@mail.nwpu.edu.cn, junwen@zju.edu.cn}
}

\maketitle

\begin{abstract}
This technical report attempts to provide efficient and solid kits addressed on the field of crowd counting, which is denoted as Crowd Counting Code Framework (C$^3$F). The contributions of C$^3$F are in three folds: 1) Some solid baseline networks are presented, which have achieved the state-of-the-arts. 2) Some flexible parameter setting strategies are provided to further promote the performance. 3) A powerful log system is developed to record the experiment process, which can enhance the reproducibility of each experiment. Our code is made publicly available at \url{https://github.com/gjy3035/C-3-Framework}. Furthermore, we also post a Chinese blog\footnote{\url{https://zhuanlan.zhihu.com/p/65650998}} to describe the details and insights of crowd counting.  
	
\end{abstract}

\section{Introduction}
\label{sec:intro}

Crowd counting is a computer vision task which treats crowd image as input, outputs corresponding crowd density map, and finally the map is summed to gain the final number of pedestrians. Recently, crowd counting has made overwhelming development with the rise of deep learning. On one hand, many large-scale datasets with human annotations are published in these years, \emph{e.g.}, \ UCF\_CC\_50~\cite{idrees2013multi}, ShangHaiTech part A and B~\cite{zhang2016single}, UCF-QNRF~\cite{idrees2018composition} and GCC~\cite{wang2019learning}. one the other hand, many CNN-based models are developed, \emph{e.g.}, MCNN~\cite{zhang2016single}, CSRNet~\cite{li2018csrnet}, SANet~\cite{cao2018scale}. However, most existing methods are running under different settings, which increases the difficulty for fair comparison. 

In this report, we are going to introduce an open-source Crowd Counting Code Framework (C$^3$ F for short) developed on pytorch\cite{pytorch}, which is an efficient and solid development kit for the crowd counting task. C$^3$F devotes to estimate a uniform and efficient code interface to conduct experiments, so that researchers and developers can benefit from it. 

\begin{figure}
	\centering
	\begin{tabular}{c}
		\includegraphics[width=0.46\textwidth]{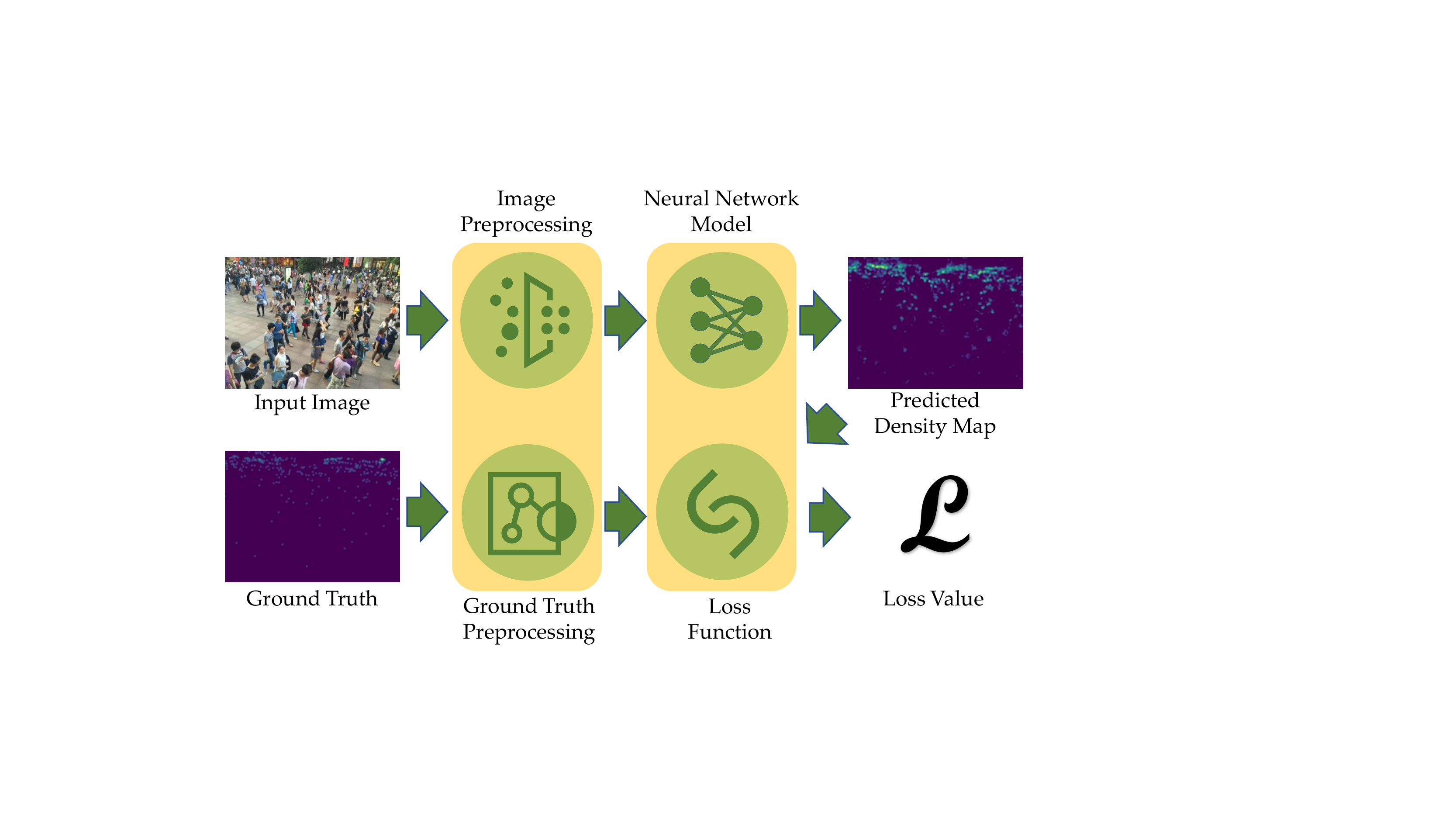}
	\end{tabular}
	\caption{Flow diagram of C$^3$F.}
	\label{fig:c3flow}
\end{figure}

Figure \ref{fig:c3flow} demonstrates the data flow and architecture of C$^3$F. It can be divided into two parts, \emph{i.e.}, data preprocessing strategy and model baseline. These two parts are introduced in Section \ref{preprocess} and Section \ref{crowdcounter}, respectively. In the final, Section \ref{conclusion} summarizes the contributions of C$^3$F.

\section{Data Preprocessing  Strategy}
\label{preprocess}

This section provides data preprocessing strategies of six mainstream datasets. They are UCF\_CC\_50 \cite{idrees2013multi}, WorldExpo'10 \cite{zhang2016data}, SHT A \cite{zhang2016single}, SHT B \cite{zhang2016single}, UCF-QNRF \cite{idrees2018composition}, and GCC \cite{wang2019learning}, as shown in Table \ref{tab:dataset}. Overall, data preprocessing strategies are constituted by  mainly two parts, including the input size and the transformation about ground truth. They are described detailedly in the following subsections.

\begin{table*}[]
	\centering
	\begin{tabular}{c!{\vrule width1.2pt}c!{\vrule width1.2pt}c}
		\Xhline{1.2pt}
		Dataset & Kernel Size               & Image Scale    \\ 
		\Xhline{1.2pt}
		UCF50\cite{idrees2013multi}   &  $15\times 15$           & keep the original height-width ratio, $\max(h, w) = 1024, \min(h, w) \% 16 = 0$ \\ \hline
		SHT A\cite{zhang2016single}   & \tabincell{c}{geometry-adaptive \\ kernels} & keep the original height-width ratio, $\max(h, w) = 1024, \min(h, w) \% 16 = 0$ \\ \hline
		SHT B\cite{zhang2016single}   & $15\times 15$            & original size: $768 \times 1024$                                                  \\ \hline
		WE\cite{zhang2016data}      & $15 \times 15$            & original size: $576 \times 720$                                                   \\ \hline
		QNRF\cite{idrees2018composition}    & $15 \times 15$            & keep the original height-width ratio, $\max(h, w) = 1024, \min(h, w) \% 16 = 0$ \\ \hline
		GCC\cite{wang2019learning}     & $15 \times 15$            & resize to $544 \times 960$                                                   \\ \Xhline{1.2pt}
	\end{tabular}
	\label{tab:dataset}
	\caption{Input image scale of different dataset. }
\end{table*}

\subsection{Input Size}
\label{subsec:is}

Operations about the input size are divided into two parts, which are image size and batch size. For image size, we restrain the height and width of input images to make sure that they are divisible by 16. This restriction guarantees some down-sampling layers (like max-pooling) could output right size as we want. More processing details are displayed in Table \ref{tab:dataset}.

As for batch size, we suggest to train through single batch size for those pre-trained models (Alexnet, VGG, ResNet, \emph{etc.}), and multiple batch size for models trained from scratch. Considering that image sizes in some dataset are different, C$^3$F advises the input tensor to be fixed in the following size when training these networks:
$$N \cdot 3 \cdot \min(h) \cdot \min(w),$$
$\min(h)$ and $\min(w)$ denote the minimum height and width of the image batch, and $N$ is the batch size. Another way is adding margin like GCC-SFCN\cite{wang2019learning}.

\subsection{Label Transformation }
\label{subsec:lt}

C$^3$F provides two operations for label transformation, including ground truth scale down-sampling and label normalization. 

The former originates from CSRNet, in which the final density maps scale is 1/8 of the original image. It firstly applies down-sampling on density maps, and then dots 64 to guarantee the sum of density map equal to the counting number. However, this operation is going to affect the PSNR and SSIM, so we do not suggest to implement this operation. C$^3$F simply stacks up-sampling layers to match the size of outputted maps and the inputted images when encountering this problem. 

Label normalization is a training trick. We find neural network could get faster convergence and lower estimation error when the density map dots a large integer value. In C$^3$F, we set this value as 100.

\section{Crowd Counting Models}
\label{crowdcounter}
In this section, we introduce some crowd counting methods modified from common classification networks (\emph{i.e.,} AlexNet\cite{krizhevsky2012imagenet}, VGG\cite{simonyan2014very}, and ResNet\cite{he2016deep}) and some representations of mainstream methods.

\subsection{AlexNet}
\begin{figure}
	\centering
	\begin{tabular}{c}
		\includegraphics[width=0.46\textwidth]{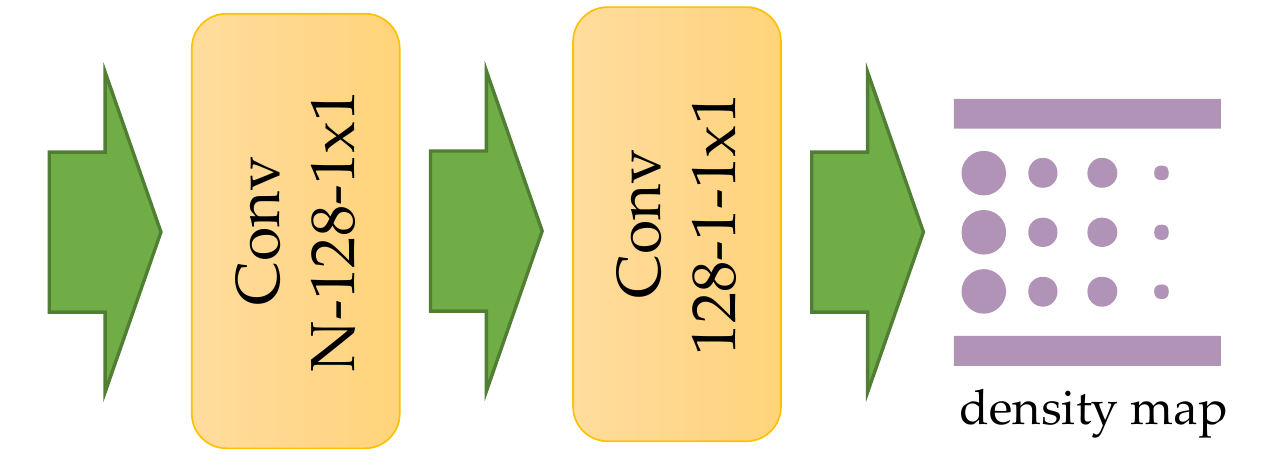}
	\end{tabular}
	\caption{Decoder Structure.}
	\label{fig:decoder}
\end{figure}

For AlexNet, we modify its padding operation in conv1 and conv2 to ensure the feature maps can be divided normaly, and only employ the network architecture before conv5 as the image feature encoder, in which the output scale is 1/16 of original image scale. The decoder is composed of two convolutional layers and an up-sampling layer, which directly regresses the final 1-channel density map, as shown in Figure \ref{fig:decoder}.

\subsection{VGG Series}
We modify VGG in two ways, both of which adopt the previous 10 convolutional layers as the encoder. The difference between them mainly lies in the decoder. VGG utilizes a simple decoder similar to AlexNet, but VGG+decoder employs another three deconvolutional layers.

\begin{table}[]
	\centering
	\begin{tabular}{|c|c|c|}
		\hline
		Method        & MAE  & MSE  \\ \hline
		VGG           & \textbf{10.3} & \textbf{16.5} \\ \hline
		VGG + decoder & 10.5 & 17.4 \\ \hline
	\end{tabular}
	\caption{Results of VGG Series.}
\end{table}

From the results of comparisons in Table 2, the performances of the above two methods are comparable with each other, but VGG+decoder produces more precise density maps. Besides, this result is similar to CSRNet (MAE: 10.6, MSE: 16.0), which also employs VGG-16 as the backbone.

\subsection{ResNet Series}

To preserve the scale of the final density maps, we change the stride of res.layer3 from 2 to 1 as the encoder, and the decoder is composed of two convolutional layers. From experimental results in Table 2, ResNet shows strong ability of image feature extraction and achieves state-of-the-art results. The best reported results of published papers are PACNN+\cite{shirevisiting}, whose MAE and MSE are 7.6 and 11.8, respectively.

\begin{table}[]
	\centering
	\begin{tabular}{|c|c|c|}
		\hline
		Method     & MAE & MSE  \\ \hline
		ResNet-50  & 7.7 & 12.6 \\ \hline
		ResNet-101 & \textbf{7.6} & 12.2 \\ \hline
		PACNN+\cite{shirevisiting}     & \textbf{7.6} & \textbf{11.8} \\ \hline
	\end{tabular}
	\caption{Results of ResNet Series.}
\end{table}

\subsection{C$^3$F Reproduction}

In this section, we reproduce some mainstream crowd counting methods, including MCNN\cite{zhang2016single}, CMTL\cite{sindagi2017cnn}, CSRNet\cite{li2018csrnet} and SANet\cite{cao2018scale}. The experimental results are presented in Table 4.

\begin{table}[]
	\centering
	\begin{tabular}{|c|c|c|}
		\hline
		Method & Original Paper & Reproduction in C$^3$F\\ \hline
		MCNN\cite{zhang2016single}   & 26.4/41.3             & \textbf{21.5}/\textbf{38.1}                         \\ \hline
		CMTL\cite{sindagi2017cnn}   & 20.0/31/1             & \textbf{14.0}/\textbf{22.3}                         \\ \hline
		CSRNet\cite{li2018csrnet} & \textbf{10.6}/\textbf{16.0}             & \textbf{10.6}/16.6                         \\ \hline
		SANet\cite{cao2018scale}  & \textbf{8.4}/\textbf{13.6}              & 12.1/19.2                         \\ \hline
	\end{tabular}
	\caption{Compare Results of some mainstream method between original paper results and representation results in C$^3$ Framework.}
\end{table}

However, we also apply some tricks on these methods. Taking MCNN for example, we do not employ single-channel but RGB images as input when reproducing it. For CMTL, C$^3$F  crops images online for more cropping regions during training. 

By the way, C$^3$F achieves the closest result to the published paper for SANet, although it is still far from its reported results.

\section{Conclusion}
\label{conclusion}
In this report, we briefly introduce a code framework C$^3$F for the crowd counting task, where the preprocessing tricks of mainstream datasets
and experimental results of modified neural networks are provided. This code framework is able to reduce the human cost in training process, and promote the academic research of crowd counting.

\paragraph{Acknowledgments}
Throughout the project, many developers have provided strong supports. Especially, thanks Xin Zeng\footnote{\url{https://https//github.com/wwwzxoe303com}} and Shuo Bai\footnote{\url{https://github.com/PetitBai}} for checking and testing the project code,  thanks Google Colab for providing free experimental resources. Besides, Some code and design logic of C$^3$F reference are borrowed from following repositories/projects/code: py-faster-rcnn\footnote{\url{https://github.com/rbgirshick/py-faster-rcnn}}, pytorch-semantic-segmentation\footnote{\url{https://github.com/zijundeng/pytorch-semantic-segmentation}}, CSRNet-pytorch\footnote{\url{https://github.com/leeyeehoo/CSRNet-pytorch}}, SANet\_implementation\footnote{\url{https://github.com/BIGKnight/SANet_implementation}}, enet.pytorch\footnote{\url{https://github.com/gjy3035/enet.pytorch}}, GCC-SFCN\footnote{\url{https://github.com/gjy3035/GCC-SFCN}}, and PCC-Net\footnote{\url{https://github.com/gjy3035/PCC-Net}}. Benefiting from these excellent open source codes, we are able to complete the C$^3$F project.

{\small
	\bibliographystyle{ieee}
	\bibliography{egbib}
}

\end{document}